\documentclass[conference]{IEEEtran}

\IEEEoverridecommandlockouts
\usepackage{amsmath,amssymb,amsfonts}
\usepackage{mathtools}
\usepackage{algorithmic}
\usepackage{textcomp}
\usepackage{xcolor}
\usepackage{array}
\usepackage{times}
\usepackage{url}
\usepackage{latexsym}
\usepackage{mathtools}
\usepackage{amsfonts}
\usepackage{tikz}
\usepackage{tikz-dependency}
\usepackage{multirow}
\usepackage{caption}
\usepackage{subfigure}
\usepackage{floatrow}
\newcolumntype{P}[1]{>{\centering\arraybackslash}p{#1}}
\newcolumntype{M}[1]{>{\centering\arraybackslash}m{#1}}
\def\BibTeX{{\rm B\kern-.05em{\sc i\kern-.025em b}\kern-.08em
    T\kern-.1667em\lower.7ex\hbox{E}\kern-.125emX}}
\begin{document}

\title{ Codeswitched Sentence Creation using Dependency Parsing}

\makeatletter
\newcommand{\linebreakand}{%
  \end{@IEEEauthorhalign}
  \hfill\mbox{}\par
  \mbox{}\hfill\begin{@IEEEauthorhalign}
}
\makeatother

\author{\IEEEauthorblockN{Dhruval Jain}
\IEEEauthorblockA{\textit{On-Device AI} \\
\textit{Samsung R \& D Institute}\\
Bangalore, India \\
dhruval.jain@samsung.com}
\and
\IEEEauthorblockN{Arun D Prabhu * \thanks{* Equal Contribution} }
\IEEEauthorblockA{\textit{On-Device AI} \\
\textit{Samsung R \& D Institute}\\
Bangalore, India \\
arun.prabhu@samsung.com}
\and
\IEEEauthorblockN{Shubham Vatsal * }
\IEEEauthorblockA{\textit{On-Device AI} \\
\textit{Samsung R \& D Institute}\\
Bangalore, India \\
shubham.v30@samsung.com}
\linebreakand
\IEEEauthorblockN{Gopi Ramena}
\IEEEauthorblockA{\textit{On-Device AI} \\
\textit{Samsung R \& D Institute}\\
Bangalore, India \\
gopi.ramena@samsung.com}
\and
\IEEEauthorblockN{Naresh Purre}
\IEEEauthorblockA{\textit{On-Device AI} \\
\textit{Samsung R \& D Institute}\\
Bangalore, India \\
naresh.purre@samsung.com }

}

\maketitle

\begin{abstract}
 Codeswitching has become one of the most common occurrences across multilingual speakers of the world, especially in countries like India which encompasses around 23 official languages with the number of bilingual speakers being around 300 million. The scarcity of Codeswitched data becomes a bottleneck in the exploration of this domain with respect to various Natural Language Processing (NLP) tasks. We thus present a novel algorithm which harnesses the syntactic structure of English grammar to develop grammatically sensible Codeswitched versions of English-Hindi, English-Marathi and English-Kannada data. Apart from maintaining the grammatical sanity to a great extent, our methodology also guarantees abundant generation of data from a minuscule snapshot of given data. We use multiple datasets to showcase the capabilities of our algorithm while at the same time we assess the quality of generated Codeswitched data using some qualitative metrics along with providing baseline results for couple of NLP tasks.
\end{abstract}


\section{Introduction}
\label{sec:intro}
Codeswitching can be defined as a norm among multilingual or bilingual societies where speakers mix multiple languages or codes while conversing. Codeswitching is one of the most common linguistic behaviours found amongst the people of countries like India which is home to several hundreds of languages. There are many social theories which provide a rationale behind the existence of such behaviours. For example, Markedness model \cite{myers1998codes} suggests that a person consciously evaluates the alternation between codes depending on the social forces in their community and decides whether to adopt or reject the normative model. According to Communication Accommodation Theory \cite{giles2016communication}, the alternation between codes helps a person in emphasizing or minimizing social differences with other people in the conversation. 

Apart from the social theories which present different logical explanations for usage of Codeswitching under various circumstantial factors, there are many linguistic theories which talk about the grammatical intricacies of Codeswitching. One of the most popular theories is Poplack’s Constraint-based model \cite{poplack2000sometimes}. This model lists down two constraints associated with Codeswitching. The first constraint which is the Equivalence constraint states that Codeswitching can only happen between surface structures of languages such that grammar properties of all the languages are satisfied. The second constraint, Free Morpheme constraint prohibits Codeswitching between a bound morpheme and a free morpheme. Although being one of the most widely accepted theories, there have been many criticisms associated with Poplack's Constraint model. For example, let us consider a sentence ``Mary handed the keys ek larke ko'' (``Mary handed the keys to a boy''). The phrase \emph{ek larke ko} is translated as \emph{a boy to}, making it grammatically incorrect in English, but still is a valid sentence in English-Hindi Codeswitching regardless of the constraints postulated by Poplack's model. Similar to Poplack's model, there are many other theories or models which have come up with their own way of defining Codeswitching grammar but all of them face criticisms in some or the other way. Our proposed work is inspired from Matrix Language Frame model \cite{myers1997duelling,myers2002contact} which is one of the latest and widely accepted theories in the field Codeswitching.

There are four types of Codeswitching. They are Intersentential Switching, Intrasentential Switching, Tag Switching and Intraword Switching. As evident from the name itself, Intersentential Switching focuses on switching between two sentences, Intrasentential Switching talks about switching within a sentence, Tag Switching concentrates on word based switching and Intraword Switching discusses about switching within a word. Most of the research revolves around Intrasentential Switching and we have also focused on the same. Some of the examples of Intrasentential Switching can be found in Table \ref{tab:Intra}.

\begin{table*}
\caption{Examples of Intrasentential Switching}
\begin{center}
\begin{tabular}{|c|c|}
\hline
\textbf{Languages} &  \textbf{Utterance}\\
\hline
English & Tommy likes swimming but he has not dived in years. \\
\hline
English - Hindi & Tommy ko swimming pasand hai but he has not dived in years \\
\hline
English - Kannada & Tommy ge swimming ishta but he has not dived in years \\
\hline
English - Marathi &  Tommila swimming awadte but he has not dived in years.  \\
\hline
\end{tabular}
\end{center}
\label{tab:Intra}
\end{table*}

With the advent of social media and artificially intelligent assistants like Siri, Bixby etc. the focus on conversational data especially the one containing Codeswitched utterances has gained significant momentum. But with the paucity of good quality Codeswitched data, most of the research gets restricted to limited scope. In our work, we propose an algorithm which makes use of structural and functional features of English grammar as defined by linguistic typology \cite{song2014linguistic} in order to produce Codeswitched data without losing the true intent or sense of the given data. Our algorithm is based on extraction of independent clauses and adjuncts present in a given sentence. These independent clauses help us in detecting the switch points in a way which ensures that grammatical dependencies are not corrupted leading to syntactic inconsistencies across languages. Also, most of the previous works of generating Codeswitched data incorporates a lot of manual effort ranging from annotations to grammar based sanity checks which our algorithm aims to reduce significantly.

The rest of the paper is organised in the following way. Section \ref{sec:Related Works} talks about related works. We elucidate the working of our algorithm in section \ref{sec:Methodology}. Section \ref{sec:Experiments} concentrates on the experiments we conducted and section \ref{sec:Results} presents the corresponding results we achieved. The final section \ref{sec:Conclusion} takes into consideration the future improvements which can be further incorporated.

\section{Related Work}
\label{sec:Related Works}
We already discussed some of the social theories regarding the importance of Codeswitching from the perspective of individual human beings. But the significance of Codeswitching from information technology point of view is even more appealing. Abundance of Codeswitched data on social media shows the need of in depth analysis of this domain in order to have clearer picture about user opinions or sentiments. Healthcare systems and education platforms if equipped with Codeswitched communication abilities could serve their customers in a much more personalised manner. In the era of virtual assistants where conversational data holds  a special place, an augmentation of Codeswitching feature can really change the market dynamics of such artificially intelligent products. Generation of abundant Codeswitched data is the first step towards conducting any noteworthy research in this domain and thereafter having a wider impact.

Many works have been presented in the domain of Codeswitching with respect to various NLP tasks. \cite{joshi2016towards} presents one of the state of the art works on Sentiment Analysis for Codeswitched data. But a closer observation on the quality of data reveals that it is monolingually dominant with minimal switch points in it. \cite{dhar2018enabling} presents it’s work on Machine Translation for Codeswitched data. The paper mentions difficulty in creation of altogether new translation system because of lack of good quality data. \cite{chandu2018language} talks about Language Modelling of Codeswitched data. The data used in this paper was collected from Hinglish blogging websites. While discussing about the challenges they faced, they mention about the lack of robustness of model as the data used was restricted to certain blogging websites. The results corresponding to these NLP tasks can be significantly improved if we develop an automated way to convert the existing monolingual resources into a Codeswitched version for a given target language which is our main focus of this paper. 

There have been many previous works which have concentrated on accumulation of Codeswitched data. \cite{chakravarthi2020corpus} creates a Codeswitched data for Tamil-English from YouTube comments for Sentiment Analysis. \cite{raghavi2015answer} creates a Question Answering Codeswitched dataset from a version of \emph{Who wants to be a billionaire?} TV show. \cite{khanuja2020new} uses Hindi movie data to create a Codeswitched version for Natural Language Inference. \cite{swami1805corpus} presents its Twitter based Hindi-English Codeswitched data for Sarcasm Detection. \cite{bohra2018dataset} discusses about Hate Speech Detection task using Hindi-English Codeswitched data gathered from Twitter. But all of these works involve a lot of dependency on resources like Twitter, YouTube etc or involve manual effort to filter wrongly identified Codeswitched data and improve quality of data. Our work provides an automated way to create Codeswitched data from existing resources. For example, we can use existing Sentiment Analysis or Question Answering datasets which are already annotated and use our concept of independent clause and adjunct extraction to get a superior version of Codeswitched data for native languages like Hindi, Kannada, Marathi with a given foreign language like English.
\begin{figure*}[t!]
\RawFloats
\centering
\includegraphics[scale=0.45]{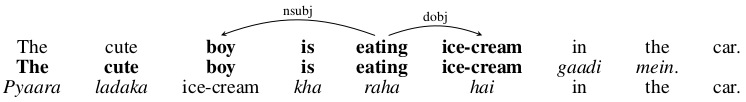}
{\caption*{Case 1}}
\vspace{0.2cm}
\includegraphics[scale=0.45]{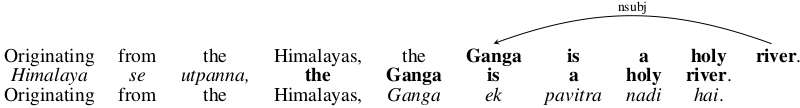}
{\caption*{Case 2}}
\vspace{0.2cm}
\includegraphics[scale=0.45]{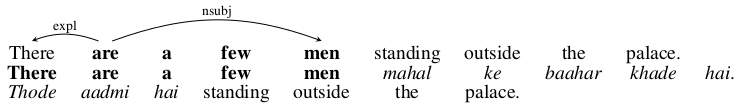}
{\caption*{Case 3}}
\vspace{0.2cm}
\includegraphics[scale=0.45]{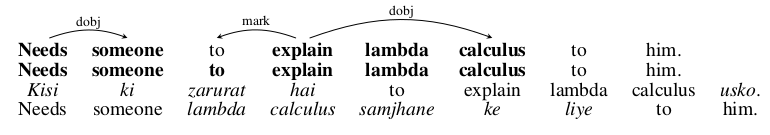}
{\caption*{Case 4}}
\caption{Examples for various Cases listed for Segment Level Analysis.}
\label{fig:case}
\end{figure*}

Some prior studies have opted ways to convert existing datasets to Codeswitched version. \cite{banerjee2018dataset} converts DSTC2 restaurant reservation dataset to Codeswitched versions of Hindi-English, Bengali-English, Gujarati-English and Tamil-English. There were two issues with the methodology adopted in this work. First, the placeholder technique used in this paper is not scalable and cannot be used for other datasets. Second, this entire dataset construction involved many annotators and evaluators, thus inculcating a lot of human interference. \cite{DBLP:journals/corr/abs-2005-00458}, though aims at generating synthetic Codeswitched data using generative adversarial training but inculcates a dependency on some amount of already existing Codeswitched data. Similarly, \cite{samanta2019deep} uses variational autoencoder architecture to synthesize Codeswitched data but again needs Codeswitched training set to achieve its goals.

\section{Methodology}
\label{sec:Methodology}
The reason behind switching of language at a particular point in a sentence is quite subjective. It may depend on person's fluency in the language or the volubility of these foreign terms in the person's community. Many theories including Poplack's Equivalence Constraint, Government Binding theory \cite{chomsky1993lectures} etc. have tried to investigate grammatical correctness of Codeswitched sentences at switch points. However, due to variations in constituency grammar across languages, there are multiple instances where these theories fail. Further, these variations increase as switch happens between languages following different word-order, i.e, English following the order subject-verb-object (SVO), Indian languages like Hindi, Kannada, Marathi, etc having subject-object-verb (SOV) order. Our work is strongly aligned with Matrix Language Frame model \cite{myers1997duelling,myers2002contact} where constituents from one language (embedding langauge) are borrowed and incorporated at acceptable points into other language (matrix language) which defines the grammatical structure for the entire Codeswitched sentence. It is quite clear that defining universal grammar at switch points in case of switching between SVO and SOV is controversial. To address this, we analyse sentences in English at phrasal and clausal levels. Independent clauses can be translated to the native language (Hindi, Kannada or Marathi) ensuring grammatical correctness with respect to both the languages. Additionally, some entities like nouns can be embedded back into this phrase after translation for increasing the frequency of switches. Hence, this way we can effectively convert English sentences into Codeswitched sentences restoring all the semantic aspects of the original sentence. In the following sub-sections, we describe our method for breaking sentences into independent clauses and adjuncts and correspondingly translating them.

\begin{figure*}[t!]
\RawFloats
\centering
\includegraphics[scale=0.33]{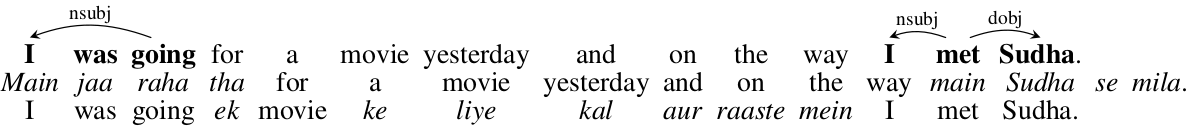}
{\caption*{(a) Proper Translation}}
\vspace{0.2cm}
\includegraphics[scale=0.42]{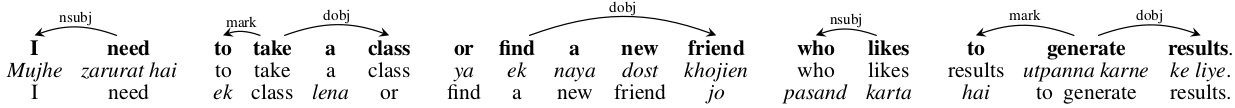}
{\caption*{(b) Improper Translation}}
\vspace{0.2cm}
\includegraphics[scale=0.38]{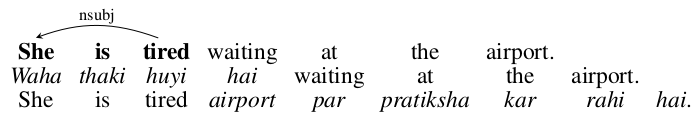}
{\caption*{(c) Improper Translation}}
\caption{Translation of various Examples}
\label{fig:trans}
\end{figure*}

\subsection{Segment Level Analysis}
A segment can be defined as a part of sentence which can be independently considered for translation. Independent clauses contain the subject and verb, so they act as the smallest unit within the sentence that makes complete sense. It is highly probable that bilingual speakers switch languages at the interface of independent clause and the adjunct. Though, some words can be borrowed from the foreign language (English) as well. For example, consider the sentence \emph{The boy is playing in the garden}. Independent clause here is \emph{The boy is playing} whereas \emph{in the garden} serves as the supporting adjunct. We can consider both independent clauses as well as adjuncts as segments while considering to translate them to one of the native languages in order to generate Codeswitched data.

Dependency parsing represents semantic relations between words in a sentence. Dependencies are triplets containing name of the relation, parent and dependent (child). They can be represented in the form of a graph with each edge representing the relation from a parent node to a child node. We use Stanford Dependency Parsing \cite{de2006generating} to extract relation between heads and dependents of the sentence. We focus on relations like nominal subject (nsubj) and direct object (dobj). The tentative clause boundaries are decided by the positions of dependents governed by the relations nsubj and dobj. We discuss the following cases of identifying various segments appropriately. The cases have been illustrated in Figure \ref{fig:case} where we have taken all examples in for English-Hindi. The same set of rules can be extended for English-Kannada and English-Marathi as well.

\subsubsection{Case 1 : (subject, head, object)}
Dependents governed by nsubj (subject) and dobj (object) have a common head and their occurrence in the sentence follows the order : subject, head and object. The continuous segment starting from subject till the object forms an independent clause. For maintaining, semantic information, we traverse the parse tree of the sentence and modify the start of this clause as the start of the noun phrase which contains the subject. For example, in the sentence shown in Figure \ref{fig:case}, Case 1, \emph{boy is eating ice-cream} forms an independent clause, and we extend the start of this clause to incorporate the noun phrase \emph{The cute boy} which contains the subject, \emph{boy}. The result of these rules finally give us our segment \emph{The cute boy is eating ice-cream}. An important thing to note here is that the remaining part of sentence \emph{in the car} serves as an adjunct and hence can also be considered as a segment. The permutation of different segments found in Figure \ref{fig:case}, Case 1 is further shown to produce different Codeswitched versions of original sentence.

\subsubsection{Case 2 : (subject, head)}
Object is absent and subject occurs before head in the sentence. Here, we modify the start of the clause with respect to the starting position of noun phrase containing the subject. The head may not always be a verb, it can be a noun or an adjective in case copular verbs are present. The end of the clause is extended to include adjectival or adverbial phrase following the head. Consider the sentence shown in Figure \ref{fig:case}, Case 2, \emph{Ganga is a holy river} contains subject before head and the noun phrase \emph{the Ganga} contains the subject. Hence, following the above discussed rule we recognize \emph{the Ganga is a holy river} as a segment.

\subsubsection{Case 3 : (head, subject)}
It may also happen that subject follows head in the sentence. In this case, we search for the first occurrence of the head's dependent and the start of clause is modified with that position. For example, in Figure \ref{fig:case}, Case 3, \emph{are a few men} represents an independent clause, and we extend the start of this clause to incorporate the dependent of head \emph{are} which is \emph{There} to get our segment. Again just like Case 1, the remaining part of our sentence forms an adjunct and hence becomes a segment. Different independent translations of segments finally give us different Codeswitched versions for the given sentence.

\subsubsection{Case 4 : (head, object)}
In cases where subject is absent, object follows its head in the sentence. Here also, we extend the start of the phrase to include the dependents of the head. This part of the sentence acts like an adjunct and adds information to the main clause, if present elsewhere in the sentence. An example is shown in \ref{fig:case}, Case 4. Here, \emph{explain lambda calculus} is the phrase where subject is absent and object follows its head. This phrase is extended to include the dependent of head which is \emph{to} to give us our segment \emph{to explain lambda calculus}. In the same sentence, we see that another phrase \emph{Needs someone} is present which has object following its head. But since in this case, there are no dependents of the head \emph{Needs}, hence the phrase \emph{Needs someone} itself qualifies for a segment.

\begin{table*}
\caption{ Original Dataset Statistics}
\begin{center}
\begin{tabular}{|c|c|c| c | c | c | c|}
\hline
\textbf{Metrics} & \textbf{DSTC2} &  \textbf{STS-Test}\\
\hline
  Number of Unique Utterances & 6733 & 498\\
 \hline
 Average Length & 7.8 & 12.63 \\
\hline
Total Vocabulary Size & 1229 & 2469 \\
\hline
\end{tabular}
\end{center}
\label{tab:ODS}
\end{table*}

\subsection{Translation}
Multiple phrases and clauses can be segmented based on the cases listed above. Each segment can be translated to the native language independently. To reduce manual effort, we use Google Translate API for translation of the segments obtained from the above cases. Segments obtained as adjuncts lack contextual information, and their translations may not fit grammatically with the other segments in the sentence. For example, In Figure \ref{fig:trans} (b), \emph{or find a new friend} should be translated to \textit{ya ek naya dost khojne ki} and similarly, in Figure \ref{fig:trans} (c), \emph{waiting at the airport} should be translated to \textit{airport par pratiksha karte huye} and hence grammatical compatibility with rest of the sentence will be lost. But in Figure \ref{fig:trans} (a), translation of \emph{for a movie yesterday and on the way} fits correctly with respect to the other segments. We can avoid this by translating only independent clause and keeping adjuncts and other phrases intact. But, this reduces the number of Codeswitched sentences generated.

\cite{samanta2019deep} have introduced variational autoencoders for generating synthetic Codeswitched sentences. They showed a significant drop in perplexity, but however, didn't comment on grammatical correctness. In our work also, we maintain the grammatical sanity in a way which does not tamper with the true sense or intent of a given sentence, but we cannot guarantee grammatical perfection in the manner of real world Codeswitched utterances.  \cite{pratapa} presented a method to generate grammatically correct Codeswitched English-Spanish sentences using Equivalence Constraint theory. As discussed in section \ref{sec:intro}, this theory restricts certain switches that are quite common in English-Hindi Codeswitching and hence lacks logical credibility for universally accepted Codeswitching grammar. Our method discussed above breaks sentence into segments across which language can be switched.

\section{Experiments}
\label{sec:Experiments}

We perform extensive evaluation of our approach on Codeswitched versions of datasets for multiple NLP tasks. We describe the evaluation metrics to calculate the complexity of our Codeswitched datasets, the tasks and datasets used and our results on these tasks in the following subsections. 

\subsection{Evaluation Metrics}
\label{subsec:CEM}
An evaluation metric is a necessity to measure the overall capabilities of any system. In case of a dataset, we measure the quality of the dataset using some metrics. We use the following metrics to measure the amount of Codeswitching obtained in our synthesized datasets: 
\subsubsection{Code-Mixing Index (CMI)}
\cite{gamback2014measuring} proposed CMI which takes into account the language distribution and switching between them to quantify the amount of Codeswitching. CMI per utterance can be calculated by : 
\begin{equation}
C_{u}(x) = \begin{cases}
 \frac{N(x) - max(t_{L_{i})(x)}} {2 * N(x)} & N(x) > 0 \\
0 & N(x) = 0
\end{cases}
\end{equation}

where N(x) are the language independent tokens in x, $L_{i} \in \mathbb{L}$ is the set of all languages in the corpus and $t_{L_i}$ is the number of tokens of $L_{i}$ in x.  The Codeswitching for the entire corpus is then calculated by: 
\begin{equation}
C_{avg} = \frac{1}{U} \sum^{U}_{x=1}C_{u}(x)
\end{equation}

\subsubsection{I-index (I \textsubscript{index})}
Switch-points are points within a sentence where the languages of the words on the two sides are different. I-index introduced by \cite{guzman2016simple} is number of switch points in the sentence divided by total number of word boundaries in the sentence. Thus, if there are $s$ number of switch points in a sentence of length $l$, then I-index can be calculated by
\begin{equation}
I_{index} = \frac{s}{l -1}
\end{equation}

\begin{table*}
\caption{Codeswitching Metrics}
\begin{center}
\begin{tabular}{|c|c|c| c | c | c | c|}
\hline
\textbf{Metrics} & \multicolumn{3}{|c|}{\textbf{DSTC2}} &  \multicolumn{3}{|c|}{\textbf{STS-Test}}\\
\hline
  & Eng - Hin & Eng - Kan & Eng - Mar &  Eng - Hin & Eng - Kan & Eng - Mar \\
\hline
C\textsubscript{avg} & 0.548 & 0.528  & 0.578 & 0.719 & 0.752 & 0.745  \\
\hline
I \textsubscript{index}  & 0.157 & 0.147 & 0.152 & 0.222 & 0.212  & 0.216 \\
\hline
\end{tabular}
\end{center}
\label{tab:CM}
\end{table*}

\begin{table*}
\caption{Codeswitched Dataset Statistics}
\begin{center}
\begin{tabular}{|c|c|c|c|c|c|c|}
\hline
\textbf{Statistical Metrics} &\multicolumn{3}{|c|}{\textbf{STS-Test}}
&\multicolumn{3}{|c|}{\textbf{DSTC2}}\\
\hline
  & Eng - Hin & Eng - Kan & Eng - Mar &
  Eng - Hin & Eng - Kan & Eng - Mar\\
\hline
English Vocabulary Size & 1078 & 998  & 995 & 793 & 761 & 777 \\ 
\hline
Native Vocabulary Size & 1250 & 1703  & 1559 & 1392 & 1046 & 1002\\
\hline 
Others Vocabulary Size & 207 & 203  & 208 & 512 & 515 & 513 \\
\hline
Codeswitched Utterances & 4942 & 4875  & 4906 & 6151 & 6258 & 6235\\
\hline
English Utterances & 3 & 9  & 7 & 34 & 51 & 44 \\
\hline
Native Utterances & 55 & 116  & 87 & 548 & 424 & 454 \\
\hline
Total Unique Utterances & 5000 & 5000  & 5000 & 6733 & 6733 & 6733 \\
\hline
\end{tabular}
\end{center}
\label{tab:DS}
\end{table*}

\subsection{Datasets}
In this paper, we use our methodology to create Codeswitched versions for two datasets and correspondingly we focus on two NLP tasks on these two generated datasets. The Codeswitching metrics along with data statistics for these two synthesized datasets are provided in section \ref{sec:Results}. We also provide the outcomes of these NLP tasks in the section \ref{sec:Results}. The original statistics for these two datasets are shown in Table \ref{tab:ODS}

\subsubsection{Dialog State Tracking Challenge 2 (DSTC2) Dataset}
DSTC \cite{williams2014dialog} is a research challenge focused on improving the state of the art in tracking the state of spoken dialog systems. DSTC2 dataset consists of a large number of dialogs related to restaurant search. \cite{banerjee2018dataset} used a mix of in-house and crowdsourced annotators to create a Codeswitched version of the DSTC2 dataset in Hindi, Bengali, Gujarati and Tamil. The limitations of this work have been already discussed in section \ref{sec:Related Works}. Apart from those limitations, we find that the dataset created is monolingually dominant. Moreover, depending on the person, different linguistic units of the native language (Hindi) can be translated to the foreign language (English). Thus, a language model trained on a dataset which is predominantly romanticised in one of the native languages may fail in the scenarios when the amount of English phrases increase. For every utterance in the dataset, we extract the independent segments and translate those segments into one of our targeted native languages which leads to maximum CMI. The test and train divisions are inherently present in DSTC2 dataset and we used the same.

\subsubsection{STS-Test}
STS-Test \cite{go2009twitter} is an evaluation dataset for Twitter Sentiment Analysis. But our data generation method is dependent on the dependency parser which does not perform well on the Twitter dataset since the dataset does not adhere to proper grammar rules and contains a lot of spelling errors. So, we clean the dataset by removing all mentions ('@' followed by Twitter handle of the user to tag), URLs and hashtags.  The spellings of the words were manually checked and corrected. We split the dataset of 498 into two parts for training and testing of sizes 398 and 100 respectively. Independent segments were then extracted from each tweet of both the datasets and every combination of the English-Hindi, English-Kannada and English-Marathi code-switched sentences were generated. We chose 4000 tweets from the generated train dataset and 1000 tweets from the generated test dataset which had the highest CMI.




\section{Results}
\label{sec:Results}
In this section we present various Codeswitching evaluation metrics for our synthesized datasets. Apart from that we also present granular statistics for each of the created datasets. Finally, we present baseline results for couple of NLP tasks and discuss the corresponding models deployed for those tasks.

\subsection{Codeswitching Metrics}

The results for various Codeswitching evaluation metrics as discussed in section  \ref{subsec:CEM} are showcased in Table \ref{tab:CM}. As we can see, for English-Hindi DSCT2 dataset I-index and CMI are more than what has been presented in \cite{banerjee2018dataset}. As per our knowledge, since this is the first time STS-Test has been converted to a Codeswitched version, these metrics define a baseline for any similar future research.

\subsection{Dataset Statistics}

The statistics corresponding to both the datasets across the chosen three native languages are tabulated in Table \ref{tab:DS}. The \emph{Others Vocabulary Size} metric in Table \ref{tab:DS} constitutes of language independent tokens like proper nouns. It can be seen that from a small STS-Test dataset of 498 utterances, we are able to generate almost 10 times of Codeswitched data with respect to original size of monolingual data. Similarly for DSTC2 dataset, depending on the number of independent segments identified we could have generated a lot of Codeswitched data but we focused only on those Codeswitched versions of a given utterance which had the maximum CMI. A snapshot of data generated by our algorithm has also been made publicly available. \footnote{https://github.com/adp01/Code-Mixed-Dataset
}




\begin{table*}
\caption{Language Model Results}
\begin{center}
\begin{tabular}{|c|c|c|c|c|c|c|}
\hline
\textbf{Metrics} & \multicolumn{3}{|c|}{ \textbf{Seq2Seq with Attention}} & \multicolumn{3}{|c|}{ \textbf{HRED}}\\
\hline
  & Eng - Hin & Eng - Kan & Eng - Mar &  Eng - Hin & Eng - Kan & Eng - Mar \\
\hline
BLEU- 4 & 58.8 & 56.2 & 53.3 & 59.6 & 57.6  & 54.4\\
\hline
 ROGUE-1 & 67.2 & 66.1 & 64.7  & 67.9 &  66.5 & 65.2 \\
 \hline

\end{tabular}
\end{center}
\label{tab:LM}
\end{table*}
\begin{table*}
\caption{Sentiment Analysis Results}
\begin{center}
\begin{tabular}{|c|c | c |  c |c |  c |  c|}
\hline
\textbf{Metrics} & \multicolumn{3}{|c|}{ \textbf{Char- LSTM}} & \multicolumn{3}{|c|}{ \textbf{Sub-word LSTM}}\\
\hline
  & Eng - Hin & Eng - Kan & Eng - Mar &  Eng - Hin & Eng - Kan & Eng - Mar \\
\hline
Accuracy & 53.6 &  51.2 & 72.5 &  57.6 &  52.8 & 54.59 \\
\hline
\end{tabular}
\label{sentiment}
\end{center}
\label{tab:SAR}
\end{table*}

\subsection{NLP Tasks}

We undertook two NLP tasks on our created Codeswitched datasets. STS-Test since being inherently a Sentiment Analysis dataset was used for same task of analysing sentiments. We did Language Modelling for our Codeswitched version of DSTC2 dataset. The architecture used for both the tasks and their corresponding results are discussed in the following sections.
\subsubsection{Language Model}

We use Sequence-to-Sequence with Attention \cite{DBLP:journals/corr/BahdanauCB14} and Hierarchical Recurrent Encoder-Decoder (HRED) \cite{serban2016building} generation based model to evaluate our Codeswitched data. We use BLEU-4 \cite{papineni2002bleu} and ROGUE-1 \cite{lin2004rouge} evaluation metrics to analyse the performance of the models. The results are summarised in Table \ref{tab:LM}. 

\subsubsection{Sentiment Analysis Model}
We compare between two models :  Character-Level models and Subword-Level representations \cite{joshi2016towards}. For both the models, we use a 2 layered LSTM with hidden layer dimension of 100. Subword-Level representations are generated through 1-D convolutions on character inputs for a given sentence. The results are shown in Table \ref{tab:SAR}.

\section{Conclusion \& Future Work}
\label{sec:Conclusion}

Codeswitching has become one of the popular areas of research especially for multilingual communities. There have been many NLP research tasks that have been undertaken in this domain ranging from Sentiment Analysis to POS tagging but most of these works cite the problem of lack of data. We have addressed this problem in our paper by designing a methodology which can use existing resources of English data to convert it into Codeswitched versions of English-Hindi, English-Kannada and English-Marathi. The results section display the true potential of our algorithm.
This algorithm is very effective in producing annotated Codeswitched versions for various datasets corresponding to various NLP tasks where we have sentence level annotations like Sentiment Analysis, Question Answering etc. But for tasks where we have word level annotations like POS tagging or NER, though our algorithm can convert the existing dataset to Codeswitched version but individual word level labelling requires manual intervention. Extension of this algorithm which can incorporate word to word mapping and hence give us complete annotated Codeswitched version of datasets involving word level labelling can be a remarkable achievement.

 \bibliographystyle{IEEEtran}
 \bibliography{IEEEabrv,bare_conf.bib}


\end{document}